\pdfoutput=1
\documentclass[11pt]{article}
\usepackage[final]{acl}
\usepackage{times}
\usepackage{latexsym}
\usepackage[T1]{fontenc}
\usepackage[utf8]{inputenc}
\usepackage{microtype}
\usepackage{inconsolata}
\usepackage{graphicx}
\usepackage{booktabs}
\usepackage{adjustbox}
\usepackage{array}
\usepackage{multirow}
\usepackage{amsmath}
\usepackage{amssymb}
\usepackage{paralist}
\usepackage{enumitem}
\usepackage{geometry}
\usepackage{subcaption}
\usepackage{tabularx}
\usepackage{tcolorbox}
\usepackage{xcolor}
\usepackage{arydshln}
\usepackage{colortbl}
\renewcommand{\arraystretch}{1.5}
\usepackage{longtable}

\title{Fine-Grained Detection of AI-Generated Text Using Sentence-Level Segmentation}

\author{
    L. D. M. S. Sai Teja\textsuperscript{1} \enspace 
    Annepaka Yadagiri\textsuperscript{1} \enspace 
    Partha Pakray\textsuperscript{1} \\
    \textbf{Chukhu Chunka}\textsuperscript{1} \enspace
    \textbf{Mangadoddi Srikar Vardhan}\textsuperscript{1} \\
    \textsuperscript{1}National Institute of Technology Silchar \\
    \texttt{\{lekkalad\_ug\_22, annepaka22\_rs, partha, chukhu, mangadoddis\_ug\_22\}@cse.nits.ac.in} \\ [1ex] 
}

\begin{document}
\maketitle
\begin{abstract}
Generation of Artificial Intelligence (\textit{AI}) texts in important works has become a common practice that can be used to misuse and abuse AI at various levels. Traditional AI detectors often rely on document-level classification, which struggles to identify AI content in hybrid or slightly edited texts designed to avoid detection, leading to concerns about the model's efficiency, which makes it hard to distinguish between human-written and AI-generated texts. A sentence-level sequence labeling model proposed to detect transitions between human- and AI-generated text, leveraging nuanced linguistic signals overlooked by document-level classifiers. By this method, detecting and segmenting AI and human-written text within a single document at the token-level granularity is achieved. Our model combines the state-of-the-art pre-trained Transformer models, incorporating Neural Networks (\textit{NN})  and Conditional Random Fields (\textit{CRFs}). This approach extends the power of transformers to extract semantic and syntactic patterns, and the neural network component to capture enhanced sequence-level representations, thereby improving the boundary predictions by the CRF layer, which enhances sequence recognition and further identification of the partition between Human- and AI-generated texts. The evaluation is performed on two publicly available benchmark datasets containing collaborative human and AI-generated texts. Our experimental comparisons are with zero-shot detectors and the existing state-of-the-art models, along with rigorous ablation studies to justify that this approach, in particular, can accurately detect the spans of AI texts in a completely collaborative text.
All our source code and the processed datasets are available in our GitHub repository\footnote{\url{https://github.com/saitejalekkala33/GenAI\_Detect\_Sentence\_Level}}.

\end{abstract}

\begin{figure}[ht]
    \centering
    \includegraphics[width=0.5\textwidth]{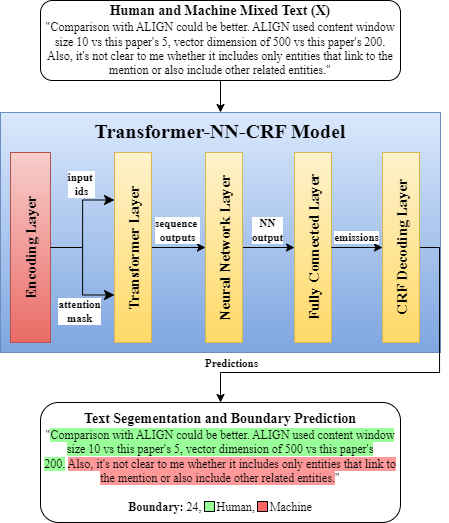}
    \caption{Text Segmentation between the Human and Machine with boundary prediction. The text in Green is Human written text, and the text in Red is Machine Generated Text.}
    \label{fig:TransformerNNCRF}
\end{figure}
\section{Introduction}

Recent advancements in Large Language Models (LLMs) \cite{chang2024survey, annepaka2025large}, such as ChatGPT\footnote{\url{https://chatgpt.com/}}, Grok\footnote{\url{https://grok.com/}}, Gemini\footnote{\url{https://gemini.google.com/}}, and DeepSeek\footnote{\url{https://chat.deepseek.com/}}, have demonstrated exceptional performance in Natural Language Processing (NLP) \cite{patwardhan2023transformers}. These models are capable of generating highly fluent and realistic human-like text, significantly enhancing applications such as text classification \cite{dogra2022complete, kowsari2019text}, sentiment analysis \cite{nasukawa2003sentiment}, machine translation \cite{stahlberg2020neural}, and question-answering systems \cite{he2024mgtbench, allam2012question}. These foundational models demonstrate significant potential in addressing a wide spectrum of NLP tasks, ranging from Natural Language Understanding (\textit{NLU}) to Natural Language Generation (\textit{NLG}), and even contributing to the development of Artificial General Intelligence (AGI) \cite{yang2024harnessing}. 
AI-plagiarism has been a growing concern in the modern world, where the distinction between human-generated and AI-generated work has become increasingly subtle. Due to the rapid growth of AI-generation tools and LLMs, academic integrity and content originality have become significant challenges. Detecting AI generation is a significant challenge for modern researchers.
Traditional plagiarism detection systems, such as Turnitin, Scribbr, etc., have fought this. Generally, detection systems such as the above systems identify plagiarism by using Machine Learning (ML) algorithms to compare the user-written text against a vast database of online content and highlight the parts that closely resemble the existing material; however, this system is often bypassed using simple techniques like paraphrasing or restructuring the sentence without altering the core meaning. This is generally due to the nature of detection methods, where general detection methods, such as perplexity-based methods, are geared towards incremental or document-level modification. This has become a significant challenge that requires great efforts from NLP researchers worldwide.

To address the above-stated problems, we propose a novel integration of a pre-trained Transformer encoder with Neural Networks and a Conditional Random Field (\textit{CRF}) layer, specifically designed for boundary-aware sentence-level authorship segmentation. Unlike prior CRF pipelines, our architecture introduces dynamic dropout and hierarchical loss-aware training, tailored to mixed-authorship detection. Unlike traditional Transformer-CRF pipelines applied to NER or POS tagging, we adapt this architecture for fine-grained authorship segmentation, incorporating sequence-aware regularization and token boundary-aware loss tailored for mixed-authorship text, a novel problem space. Our contribution lies in the hierarchical integration of authorial cues at different granularity levels, use of dynamic dropout to mitigate overfitting to local styles, and a custom-designed loss that emphasizes boundaries. To our knowledge, no existing model has applied these optimizations specifically for sentence-level segmentation in human-AI collaborative texts.

\section{Related Work} 
Based on the evolution of AI models, this mixed text data is prone to plagiarism and authenticity issues. There have been works based on the mixed text data classification by RoFT \cite{dugan2020roft} created a dynamic benchmark where users attempted to identify transition points between human and AI texts. In the follow-up RoFT-ChatGPT \cite{kushnareva2023ai} adapted SOTA detectors to locate the boundaries and compared the perplexity based approaches. The work \textit{`Towards Automatic Boundary Detection for Human-AI Collaborative Hybrid Essay in Education'} by \cite{zeng2024towards} introduced a new segmentation approach that learns distinct human and AI authorship prototypes within embedding space, while treating the boundaries as points of maximum prototype distance. SemEval 2024 shared task by \cite{wang2024m4gt}, in which many participated for the task of detecting the boundaries, in that \citet{qu2024tm} (\textit{TM-TREK}) achieved 1st place in the human-machine mixed‑text detection subtask by reframing boundary detection as token classification and leveraging ensembles of LLMs with segmentation loss. Another work, SeqXGPT \cite{wang2023seqxgpt}, introduced a sequence labeling approach using GPT-style decoder-only models as the core features.

With the rapid advancement of LLMs, collaborative writing between humans and AI has become increasingly seamless. For instance, previous work by \citet{buschek2021impact} explored the use of GPT-2 to provide phrase-level writing suggestions during email composition, aiming to support and enhance human writing. This tells about the growing relevance of detecting and understanding hybrid texts that combine contributions from both humans and AI. \citet{lee2022coauthor} extended this line of work by employing the more advanced GPT-3 model to offer sentence-level suggestions in human-AI collaborative essay writing. As this type of collaborative writing between humans and LLMs becomes more prevalent, it introduces a critical challenge for the AI text detection field: identifying AI-generated segments within jointly authored text. Addressing this issue, \citet{dugan2023real} reframed the task of detecting AI text in hybrid documents as a boundary detection problem, aiming to precisely locate the transition points between human and AI-generated text. The study specifically evaluated human ability to identify boundaries within hybrid texts containing a single transition point. Although detection accuracy improved with some level of training, overall performance remained limited, participants were only able to correctly identify the boundary in 23.4\% of the cases. Building upon the work of \citet{dugan2023real}, our research also focuses on boundary detection. However, key differences include: (1) this work explores automated methods for identifying boundaries, and (2) the hybrid texts used in our experiments contain multiple boundaries.

\begin{tcolorbox}[
    colback=blue!5!white, 
    colframe=blue!85!black,
    fonttitle=\bfseries,
    title=OUR KEY CONTRIBUTIONS:,
    boxsep=1mm, 
    left=0.5mm, 
    right=0.5mm,
    top=1mm, 
    bottom=1mm 
]
\begin{enumerate}
    \setlength{\itemsep}{0.5mm} 
    \setlength{\parskip}{0mm}   
    \setlength{\parsep}{0mm}    
    \item CRF tagged Transformer Models.
    \item Layer-wise Learning Rate Decay.
    \item Dynamic Dropout.
    \item Sequence Predictions via CRF Decoding.
    \item Xavier Initialization for Weights Stability.
    \item CRF Loss Calculation with Masking.
    \item Multiple Boundary Prediction Flexibility.
\end{enumerate}
\end{tcolorbox}

\section{Dataset Descriptions}  \label{datasets}
In this section, we used two publicly available datasets: (1) TriBERT from the paper \textit{Towards Automatic Boundary Detection for Human-AI Collaborative Hybrid Essay in Education} \cite{zeng2024towards} and (2) \textit{M4GT-Bench Task 3: Mixed Human-Machine Text Detection} \cite{wang2024m4gt}.\\
\noindent
\textbf{TriBERT}: This dataset contains various mixed author\footnote{H-Human, M-Machine} sequences such as \textit{``H-M'', ``M-H'', ``H-M-H'', ``M-H-M'', ``H-M-H-M-H'' and ``M-H-M-H-M''}, totaling six types, as seen in the Table~\ref{tab:dataset1}. \\
\textbf{M4GT}: This dataset includes only the ChatGPT and LLaMA reviews as shown in Table~\ref{tab:dataset2}. 
It follows a single pattern, \textit{``H Initiated and M Ended''}, and provides word-level boundaries indicating the exact transition point. The dataset statistics, such as train, dev, and test splits, can be seen in Table~\ref{tab:datasetsplits}. Further detailed dataset descriptions are given in Appendix~\ref{appendix_datasets}.

\begin{table}[ht]
\centering
\footnotesize
\begin{tabular}{lccc}
\toprule
\textbf{Dataset} & \textbf{Train} & \textbf{Dev} & \textbf{Test} \\
\midrule
TriBERT \cite{zeng2024towards} & 12,049 & 2,527 & 2,560 \\
M4GT \cite{wang2024m4gt} & 18,245 & 2,525 & 11,123 \\
\bottomrule
\end{tabular}
\caption{Dataset statistics from TriBERT \cite{zeng2024towards} and M4GT \cite{wang2024m4gt} showing the number of instances in train, dev, and test splits.}
\label{tab:datasetsplits}
\end{table}

\section{Proposed Methodology}
The methodology we propose integrates a hierarchical architecture that combines contextual encoding, sequential pattern modeling, and prediction to perform fine-grained sequence labeling.

\subsection{Problem Formulation}
Given a hybrid text paragraph $\mathcal{T} = \langle s_1, \ldots, s_n \rangle$, where sentences $s_i$ are human-initiated and machine-ended or mixed in \textit{HM, MH, HMH, MHM, HMHMH, and MHMHM} setting, \textit{boundary detection} identifies token-level authorship transitions. Unlike sentence-level detection, we target sub-sentence boundaries, using token sequence $\mathcal{W} = \langle w_1, \ldots, w_m \rangle$ to predict $\mathcal{Y} = \langle y_1, \ldots, y_m \rangle$, with $y_j = 1$ for AI tokens and $y_j = 0$ for human. This supports fine-grained segmentation of human-AI collaborative text.

\subsection{Model Description}
The flow of input token sequence processing begins with a pretrained Transformer-based encoder to produce rich contextual representations, and also for capturing both local and global dependencies in the overall sequence. These embeddings are then passed through a neural network layer, enabling the model to further refine token-level features by explicitly modeling sequential patterns in both forward and backward directions (a bidirectional mode). The output of the sequential layer is fed into a linear classification head that projects each token representation into a distribution over target labels. To enforce global consistency in the predicted label sequences and effectively capture dependencies among adjacent labels, the model employs a Conditional Random Field (\textit{CRF}) as the final decoding layer, and the reason for choosing the CRF is discussed in the following Section \ref{why_crfs}. During training, the model tries to minimize the negative log-likelihood of the correct label sequence under the CRF, while during inference time, it uses \textit{Viterbi} decoding to predict or output the most likely sequence of labels. While this combined-component design yields powerful representational capacity, it also increases the complexity of the model architecture and creates a chance for overfitting. \\
\noindent \textbf{Model Optimization:} To stabilize the above concerns, several optimization techniques are incorporated into the model, those are 1) including \textit{Layer-wise Learning Rate Decay} (\textit{LLRD}) to stabilize fine-tuning across the layers, 2) \textit{Dynamic Dropout} to regularize learning by adapting to training dynamics, and 3) \textit{Xavier initialization} to ensure controlled variance propagation in the linear classification layer. This hybrid-hierarchical architecture with the optimization techniques allows the model to leverage deep contextual knowledge, sequential structure, and label inter-dependencies, making it particularly suitable for the segmentation of human and AI texts while considering the spans of the predicted labels, and it can be said that this model is well suited for the structured prediction tasks in natural language processing. A clear text segmentation and boundary prediction workflow is shown in Figure~\ref{fig:TransformerNNCRF}.

\subsection{Why CRFs?}
\label{why_crfs}
Conditional Random Fields (\textit{CRFs}) are discriminative models well suited to prediction tasks, particularly sequence labeling. Unlike Hidden Markov Models (\textit{HMMs}) and Maximum Entropy Markov Models (\textit{MEMMs}), CRFs avoid issues like label bias by modeling the conditional probability of label sequences directly.

We employ a standard linear-chain CRF layer to model inter-label dependencies. For mathematical formulation, refer \citet{lample2016neural}. For a given sequence input \( x = (x_1, x_2, \dots, x_n) \) and label sequence \( y = (y_1, y_2, \dots, y_n) \), CRF defines the probability as follows:

\begin{equation}
\label{eq:pyx_prob}
\scriptsize
    P(y \mid x) = \frac{\exp \left( \sum_{t=1}^{n} \psi_t(y_t, x, t) + \sum_{t=1}^{n-1} \phi_t(y_t, y_{t+1}, x) \right)}{Z(x)}    
\end{equation}

where: \( \psi_t(y_t, x, t) \): The score function for assigning label \( y_t \) to the \( t \)-th token in \( x \). 
\( \phi_t(y_t, y_{t+1}, x) \): The score function for the transition between labels \( y_t \) and \( y_{t+1} \). \( Z(x) \): The partition function, normalizing the probabilities:

\begin{equation}
\label{eq:zofx}
\scriptsize
Z(x) = \sum_{y' \in \mathcal{Y}(x)} \exp \left( \sum_{t=1}^{n} \psi_t(y_t, x, t) + \sum_{t=1}^{n-1} \phi_t(y_t, y_{t+1}, x) \right)
\end{equation}
CRF Score Function: The unnormalized score \( S(x, y) \) for a given sequence \( y \) is:
\begin{equation}
\label{crf_scorer}
\scriptsize
S(x, y) = \sum_{t=1}^{n} \psi_t(y_t, x, t) + \sum_{t=1}^{n-1} \phi_t(y_t, y_{t+1}, x)
\end{equation}

\textbf{CRFs vs HMMs.} Hidden Markov Models (\textit{HMMs}) represent observed data as a sequence of events, where each observation depends on a hidden state in a hidden Markov chain. Compared to CRFs, HMMs impose more significant constraints, as each state relies on a fixed set of previous hidden states, allowing them to capture local context effectively. In contrast, CRFs incorporate both local and global contexts more comprehensively using feature functions. For example, the feature function label the first word as a verb if the sequence ends with a question mark and this isn’t possible with HMMs. As a result, CRFs can be more accurate than HMMs if we define the right feature functions.

\textbf{CRFs vs MEMMs.} MEMM combines HMM with the Maximum Entropy (log-linear) classifiers. Unlike generative models like HMMs, MEMMs are discriminative, but they suffer from the `label bias problem', where states with fewer outgoing transitions may ignore observations. CRFs addresses them by modeling joint probability distributions to capture dependencies between labels while mitigating bias. This makes CRFs particularly suitable for sentence-level classification tasks.

\section{Experiments}
\textbf{Experimental Setup:} We conducted all our experiments on an Amazon Web Services (AWS) cloud server. In the EC2 instance, we initiated an instance for Accelerated Computing. The specifications are \textbf{g6e.xlarge} instance, which provides 3rd generation AMD EPYC processors (AMD EPYC 7R13), with a \textbf{NVIDIA L40S Tensor Core GPU with 48 GB GPU memory}, and 4x vCPU with 150 GiB memory, and our OS type is Ubuntu Server 24.04 LTS (HVM).

We conducted our experiments by combining CRF layer to the sequential models starting from the Neural Networks to the Transformer models, where the neural network models we have taken are Convolutional Neural Network (\textit{CNN}), Recurrent Neural Network (\textit{RNN}), Long Short-Term Memory (\textit{LSTM}), Bidirectional LSTM (\textit{BiLSTM}), and Bidirectional Gated Recurrent Unit (BiGRU), and the transformer models we have taken are the BERT \cite{devlin2018bert}, DistilBERT \cite{sanh2019distilbert}, RoBERTa \cite{liu2019roberta}, DeBERTa \cite{he2020deberta} and ModernBERT \cite{warner2024smarter}, and our proposed architecture is the combination of a Transformer, a Neural Network and a CRF layer. We conducted several combinations of the above models. As we mentioned, the emergence of CRF with the prior models HMMs and MEMMs, we experimented with these by replacing the CRF with HMM and MEMM in the best-performing model with CRF. All our experiments can be seen in the Table~\ref{tab:all_experiment_included}, and the hyperparameters can be seen in Table~\ref{tab:hyperparameters}.

\begin{table}[ht]
\centering
\renewcommand{\arraystretch}{0.9} 
\setlength{\tabcolsep}{6pt}       
\begin{tabular}{lc}
\toprule
\textbf{Hyperparameter} & \textbf{Value} \\
\midrule
batch\_size & 32 \\
epochs & 3 \\
gradient\_clip & 1.0 \\
hidden\_dim & 512 \\
num\_layers & 3 \\
num\_labels & 2 \\
weight\_decay & 1e-2 \\
max\_len & 512 tokens \\
learning\_rate & 1e-6, 5e-6, 1e-5, 1e-4 \\
\bottomrule
\end{tabular}
\caption{Hyperparameters}
\label{tab:hyperparameters}
\end{table}

\begin{table}[h]
\centering
\resizebox{1.0\linewidth}{!}{%
\begin{tabular}{p{4cm}p{11cm}}
\toprule
\textbf{Type} & \textbf{Models} \\
\midrule
Neural Network + CRF & CNN,\quad RNN,\quad LSTM,\quad BiLSTM,\quad \textbf{BiGRU}\\
\hdashline
Transformer + CRF & BERT,\quad DistilBERT ,\quad RoBERTa,\quad \textbf{DeBERTa},\quad ModernBERT\\
\hdashline
Transformer + NN + CRF & DeBERTa + CNN,\quad DeBERTa + RNN,\quad DeBERTa + LSTM,\quad DeBERTa + BiLSTM,\quad \textbf{DeBERTa + BiGRU},\quad BERT + BiGRU,\quad DistilBERT + BiGRU,\quad RoBERTa + BiGRU,\quad ModernBERT + BiGRU \\
\hdashline
T + NN + HMM/MEMM & DeBERTa + BiGRU + HMM,\quad DeBERTa + BiGRU + MEMM \\
\bottomrule
\end{tabular}%
}
\caption{All our experimental combinations with the Neural Networks and the transformer models, the best performing models in each setting highlighted in Bold.}
\label{tab:all_experiment_included}
\end{table}

\subsection{Comparisons With SOTA Techniques}
We have compared our approach's best performing model with the zero-shot training free models and others with the prior models that were experimented on these datasets previously and lastly with the HMM and MEMM models. In total they are:
\begin{enumerate}
    \item \textbf{Zero-Shot methods for both M4GT and TriBERT datasets}
    \begin{enumerate}
        \item FastDetectGPT - `\textit{falcon-7b-instruct}'
        \item FastDetectGPT - `\textit{gpt-neo-2.7b}'
        \item Glimpse - `\textit{davinci-002}'
        \item Glimpse - `\textit{babbage-002}'
        \item Binoculars
    \end{enumerate}
    \item \textbf{HMM \& MEMM for both M4GT and TriBERT datasets}
    \begin{enumerate}
        \item Best Performing Model\footnote{Best Performing Model - This is the model with the CRF that has best results in all metrics.} + HMM
        \item Best Performing Model + MEMM
    \end{enumerate}
    \item \textbf{For TriBERT} - \citet{zeng2024towards}
    \begin{enumerate}
        \item TriBERT (p=2) - \cite{zeng2024towards}
        \item GigaCheck (DN-DAB-DETR) - \cite{tolstykh2024gigacheck}
    \end{enumerate}
    \item \textbf{For M4GT} - \citet{wang2024m4gt}
    \begin{enumerate}
        \item Longformer - \cite{wang2024m4gt}
        \item DeBERTa-V3 - \cite{wang2024m4gt}
        \item TM-TREK - \cite{qu2024tm}
        \item AIpom - \cite{qu2024tm}
        \item USTC-BUPT - \cite{guo2024ustc}
    \end{enumerate}
\end{enumerate}

\subsection{Zero-Shot Methods}
As mentioned above, we have utilized the zero-shot models for both datasets.  The zero-shot models are: 1)~\textbf{Fast-DetectGPT} \cite{bao2023fast, mitchell2023detectgpt}, with two variations: (a)~\textit{falcon-7b/falcon-7b-instruct}, where \texttt{falcon-7b} is used as the sampling model and \texttt{falcon-7b-instruct} as the scoring model, and (b)~\textit{gpt-neo-2.7b}, where the same model is used as both the sampling and scoring models. Fast-DetectGPT is built to detect AI-generated text by giving the perturbing samples and evaluating the texts likelihood under different model prompts, in which this makes the model highly adaptable when combined with different Language Models. 2)~\textbf{Glimpse} \cite{bao2023fast}, with two additional variations: (a)~\textit{davinci-002} and (b)~\textit{baggage-002}, where each model is used solely as a scoring model. Glimpse works on the principle of uncertainty-guided token perturbation, where this methods makes the model effective at identifying unnatural patterns in text. 3)~The final detector is \textbf{Binoculars} \cite{hans2401spotting}, which performs fine-grained AI detection by comparing representations between base- and instruction-tuned language models. To adapt these zero-shot methods for sentence-level segmentation, we apply them individually to each sentence using standardized prompts and aggregate the outputs.

\subsection{Existing SOTA Models}
TriBERT model that is proposed by \citet{zeng2024towards}, is a boundary detection model that uses a tri-branch structure for identifying human and AI segments in hybrid essays. GigaCheck \cite{tolstykh2024gigacheck} uses a DeNoising - Dynamic Anchor Box - DEtection TRansformer (\textit{DN-DAB-DETR}) architecture designed to detect LLM-generated text using dense attention and boundary-aware features. M4GT \cite{wang2024m4gt} benchmarked a variety of detectors, including Longformer, which uses long-range attention to identify stylistic shifts and DeBERTa-V3, a strong pretrained model adapted for segment-wise detection, TM-TREK \cite{qu2024tm}, which uses LLMs with a temporal-aware retrieval strategy, AIpom \cite{shirnin2024aipom} includes prompt engineering and contrastive learning, and last one USTC-BUPT \cite{guo2024ustc} enhances detection by aligning domain adversarial learning with LLM-derived features.

\section{Evaluation Metrics} \label{metrics}
The evaluation metrics were considered separately for each of the datasets. For the TriBERT dataset, we utilized the same metrics that were given in the paper by \citet{zeng2024towards} which is the F1 score. The authors of the paper \cite{zeng2024towards} considered two things a) $L_{topK}$ and b) $L_{Gt}$, which gives top-K boundaries detected by the model's algorithm, and the number of ground-truth boundaries respectively and the value of K is set to 3 in their analysis. The F1 score is then determined using the following formula:
\begin{equation}
    F1@K = 2 \cdot \frac{|L_{\text{top}K} \cap L_{\text{Gt}}|}{|L_{\text{top}K}| + |L_{\text{Gt}}|}
\end{equation}

For the M4GT dataset, the authors mentioned the Mean Absolute Error (MAE) of the actual boundary \( y_i \) and the predicted boundary \( \hat{y}_i \) as the main metric and \( n \) as total boundary pairs, so we have taken the same as for the comparison purpose. 

\begin{equation}
\text{MAE} = \frac{1}{n} \sum_{i=1}^{n} \left| y_i - \hat{y}_i \right|
\end{equation}

Besides these official metrics, we also computed standard evaluation metrics such as Accuracy, Precision, Recall, F1-Score, Matthews correlation coefficient (\textit{MCC}), and Cohen's Kappa score. These metrics are computed based on the token-level prediction, as the spans or boundaries are determined by the binary labels (0-`human' or 1-`AI') of token sequences. Each token in the dataset is labeled as 0 or 1 for the CRF computation during the training, and during the inference, these labels will be predicted using the Viterbi algorithm of the CRF, such that these are computed whether the prediction is 0 or 1.
The Viterbi algorithm finds the sequence \(\hat{y}\) with the maximum score as in Equation~\ref{max_score} and the recursive formulation used in the Viterbi algorithm is showed in the below Equation~\ref{recurscive formulation}.
\begin{equation}
\label{max_score}
\hat{y} = \arg \max_{y \in \mathcal{Y}(x)} S(x, y)
\end{equation}
\begin{equation}
\label{recurscive formulation}
\scriptsize
\delta_t(y_t) = \max_{y_{t-1}} \left[ \delta_{t-1}(y_{t-1}) + \phi_t(y_{t-1}, y_t, x) + \psi_t(y_t, x, t) \right]
\end{equation}
where s(x, y) is the CRF score function which can be seen in the Equation~\ref{crf_scorer}, \( \delta_t(y_t) \) is the maximum score of sequences ending in \( y_t \) at position \( t \).

\section{Results Analysis}

We conclude that our method overcame not only the results of the existing zero-shot methods but also the models that were previously experimented on these datasets. We hypothesize that our model’s superior performance is due to its explicit modeling of inter-sentence transitions using neural networks in between, which captures stylistic changes more effectively than token-level detectors. Zero-shot methods underperformed in comparison to our proposed model, indicating limited reliability when applied in real-world hybrid texts. On the M4GT dataset, our best-performing model (DeBERTa + BiGRU + CRF with all optimizations) achieved a MAE of 8.47, substantially outperforming the best-performing zero-shot method, which recorded an MAE of 42.37. While some prior supervised models (DeBERTa-V3) reported competitive results (MAE of 15.55), our model still demonstrates a significant margin of improvement. Conversely, in the TriBERT dataset, a few zero-shot detectors such as FastDetectGPT (\texttt{falcon-7b-instruct}) and Binoculars showed comparatively stronger performance, achieving an average $F1@K$ of 0.608 and outperforming several prior supervised baselines. Nevertheless, our proposed model consistently outperformed all baselines, including both zero-shot and prior supervised models, across varying author-mixture types. Comprehensive results for M4GT and TriBERT comparisons are provided in Tables~\ref{tab:m4gt_comparison} and~\ref{tab:tribert_comparison}, respectively.

\begin{table}[ht]
\centering
\scriptsize
\begin{tabular}{lcc}
\toprule
\textbf{Type} & \textbf{Model} & \textbf{MAE} \\
\midrule
\multirow{5}{*}{Zero-shot} 
    & Glimpse (babbage-002) & 78.84 \\
    & Glimpse (davinci-002) & 72.63 \\
    & FastDetectGPT (gpt-neo-2.7b) & 75.19 \\
    & FastDetectGPT (falcon-7b-instruct) & 48.91 \\
    & Binoculars & 42.37 \\
\hdashline
\multirow{5}{*}{Prior Works} 
    & Longformer \cite{wang2024m4gt} & 21.54 \\
    & USTC-BUPT \cite{guo2024ustc} & 17.70 \\
    & AIpom \cite{shirnin2024aipom} & 15.94 \\
    & TM-TREK \cite{qu2024tm} & 15.68 \\
    & DeBERTa-V3 \cite{wang2024m4gt} & 15.55 \\
\hdashline
\multirow{2}{*}{HMM \& MEMM} 
    & \textit{DeBERTa + BiGRU + MEMM} & 26.73 \\
    & \textit{DeBERTa + BiGRU + HMM} & 22.19 \\
\hdashline
\multirow{2}{*}{Proposed} 
    & \textbf{Ours\textsuperscript{*} - (\textit{DeBERTa + BiLSTM + CRF})} & \textbf{13.95} \\
    & \textbf{Ours\textsuperscript{*} - (\textit{DeBERTa + BiGRU + CRF})} & \textbf{8.47} \\
\bottomrule
\end{tabular}
\caption{Comparison on the M4GT \cite{wang2024m4gt} dataset with both the zero-shot methods and the prior models that utilized this dataset. The best results are marked in bold and \textsuperscript{*} denotes the model with all optimizations and best hyperparameters.}
\label{tab:m4gt_comparison}
\end{table}

\begin{table}[htbp]
    \centering
    \scriptsize
    \resizebox{1.0\linewidth}{!}{%
    \begin{tabular}{lccccc}
        \toprule
        \textbf{Type} & \textbf{Model} & \textbf{Bry=1} & \textbf{Bry=2} & \textbf{Bry=3} & \textbf{All} \\
        \midrule
        \multirow{5}{*}{Zero-shot}
        & Glimpse (babbage-002) & 0.194 & 0.256 & 0.303 & 0.251 \\
        & Glimpse (davinci-002) & 0.427 & 0.506 & 0.564 & 0.499 \\
        & FastDetectGPT (gpt-neo-2.7b) & 0.253 & 0.317 & 0.362 & 0.310 \\
        & FastDetectGPT (falcon-7b-instruct) & 0.482 & 0.553 & 0.619 & 0.551 \\
        & Binoculars & 0.517 & 0.624 & 0.683 & 0.608 \\
        \hdashline
        \multirow{2}{*}{Prior Works}
        & TriBERT (p=2) \cite{zeng2024towards} & 0.455 & 0.692 & 0.622 & 0.575 \\
        & GigaCheck \cite{tolstykh2024gigacheck} & 0.444 & 0.693 & 0.801 & 0.646 \\
        \hdashline
        \multirow{2}{*}{HMM \& MEMM}
        & \textit{DeBERTa + BiGRU + MEMM} & 0.562 & 0.717 & 0.754 & 0.677 \\
        & \textit{DeBERTa + BiGRU + HMM} & 0.593 & 0.749 & 0.787 & 0.710 \\
        \hdashline
        \multirow{2}{*}{Proposed}
        & \textbf{Ours\textsuperscript{*} - (\textit{DeBERTa + BiLSTM + CRF})} & \textbf{0.612} & \textbf{0.734} & \textbf{0.817} & \textbf{0.721} \\
        & \textbf{Ours\textsuperscript{*} - (\textit{DeBERTa + BiGRU + CRF})} & \textbf{0.695} & \textbf{0.846} & \textbf{0.878} & \textbf{0.806} \\
        \bottomrule
    \end{tabular}
    }
    \caption{Comparison on the TriBERT \cite{zeng2024towards} dataset with the zero-shot methods, the prior model that utilized this dataset and and the HMM \& MEMM models, the best ones are marked in bold and \textsuperscript{*} denotes the model with all optimizations and best hyperparameters.}
    \label{tab:tribert_comparison}
\end{table}

The evaluation metrics like Accuracy, Precision, Recall, F1-Score, MCC, and Kappa's scores are really high, scoring around 98\% and 91\% of accuracy in the TriBERT dataset and the M4GT dataset, respectively, which can be seen in the Tables~\ref{tab:master_results_AAAI} and \ref{tab:master_results_M4GT}. These results might initially suggest excellent model performance and say that the model is highly effective at distinguishing between human-written and AI-generated collaborative text. However, these traditional metrics primarily reflect token-level or word-level classification, meaning they only assess whether a single word is correctly labeled or not among the binary labels (0 and 1). In automatic boundary detection, these metrics cannot assess to capture the border structural coherence and make the correct segmentation. In contrast, the main evaluation considered by the authors involves identifying the correct segmentation of the human and AI parts, which are better and task-relevant criteria. This segmentation-based evaluation provides a clearer picture of how well the model understands distinct regions of authorship within a collaborative text, offering a more appropriate measure of performance for this work. We conclude that the Transformers model has long-range dependencies and lacks explicit span modeling. By adding Neural Network and CRF layers, we align local contextual cues with structured transitions, enabling more accurate detection of authorship shifts.

\section{Ablation Study}

We mentioned that our approach model is too complex due to its hybrid-hierarchical component architecture. To reduce the complexity while maintaining the performance good, we used techniques, 1) Layer-wise Learning Rate Decay (\textit{LLRD}), 2) Dynamic Dropout, and 3) Xavier initialization of weights. To check the usability of these techniques, we went on an ablation by including one-on-one, and finally, all techniques were added. An overview of the significance of these techniques is given below. 

\begin{enumerate}
    \item \textbf{Significance of these techniques:}
    \begin{enumerate}
        \item Xavier Initialization is used in the fully connected layer to ensure stable weight initialization.
        \item In LLRD, earlier layers of the transformer (embeddings and encoder layers) have lower learning rates, while the later encoder layers, NN, FC, and CRF layers, have progressively larger learning rates. (LR: 1e-6 $\rightarrow$ 5e-6 $\rightarrow$ 1e-5 $\rightarrow$ 1e-4)
        \item Dynamic Dropout changes the dropout rate based on the layer depth or training progression, significantly ensuring an optimal balance between regularization and learning capacity.
    \end{enumerate}
\end{enumerate}

\noindent Other than these techniques, we also mentioned the usage of HMM and MEMM instead of CRF, for that, we built the model with the same configuration as CRF model with the HMM\footnote{DBGH - DeBERTa + BiGRU + HMM} and MEMM\footnote{DBGM - DeBERTa + BiGRU + MEMM}. 

\noindent\textit{\textbf{We did this ablation for only the Best Performing model, which is the TNC-DBGC\textsuperscript{*}\footnote{TNC-DBGC: Transformer + Neural Network + CRF - DeBERTa + BiGRU + CRF} model with all the optimization techniques and the best hyperparameters.}}

\begin{table}[h!]
\centering
\scriptsize
\begin{tabular}{ll}
\toprule
\textbf{Model} & \textbf{MAE} \\
\midrule
DBGM\textsuperscript{*} & 26.73 \\
DBGH\textsuperscript{*} & 22.19 \\
\hdashline
TNC-DBGC   & 19.85 \\
+ Dynamic Dropout         & 16.42 \\
+ Xavier Initialization    & 12.76 \\
+ LLRD       & 10.03 \\
\hdashline
+ \textbf{All (TNC-DBGC\textsuperscript{*})}  & \textbf{8.47}  \\
\bottomrule
\end{tabular}
\caption{Ablation study on the M4GT dataset with the HMMs and MEMMs and the inclusion and exclusion of the optimization techniques.}
\label{tab:m4gt_ablation}
\end{table}

\begin{table}[h!]
\centering
\scriptsize
\resizebox{1.0\linewidth}{!}{%
\begin{tabular}{lcccc}
\toprule
\textbf{Model} & \textbf{Bry=1} & \textbf{Bry=2} & \textbf{Bry=3} & \textbf{All} \\
\midrule
DBGM*      & 0.562 & 0.717 & 0.754 & 0.677 \\
DBGH*      & 0.593 & 0.749 & 0.787 & 0.710 \\
\hdashline
TNC-DBGC   & 0.625 & 0.775 & 0.812 & 0.737 \\
+ Dynamic Dropout        & 0.645 & 0.798 & 0.838 & 0.760 \\
+ Xavier Initialization     & 0.670 & 0.821 & 0.867 & 0.786 \\
+ LLRD       & 0.681 & 0.835 & 0.871 & 0.795 \\
\hdashline
+ \textbf{All (TNC-DBGC\textsuperscript{*})}  & \textbf{0.695} & \textbf{0.846} & \textbf{0.878} & \textbf{0.806} \\
\bottomrule
\end{tabular}
}
\caption{Ablation study on the TriBERT dataset with the HMMs and MEMM and the inclusion and exclusion of the optimization techniques.}
\label{tab:tribert_ablation}
\end{table}

According to the above Tables~\ref{tab:m4gt_ablation} and \ref{tab:tribert_ablation}, which present a detailed ablation study on the effects of inclusion and exclusion optimization techniques alongside traditional probabilistic sequence models such as HMMs and MEMMs, it shows that the involvement of these optimization strategies plays a very crucial role in enhancing model performance. These clearly show the necessity for the model's progressive improvement in performance when optimization techniques are sequentially applied to the base TNC-DBGC model.

In the Table~\ref{tab:m4gt_ablation}, which focuses on the M4GT dataset, we can observe a significant difference in the MAE between the models TNC-DBGC and All (TNC-DBGC\textsuperscript{*}) from 19.85 to 8.47, respectively. This nearly 60\% reduction in MAE underscores the effectiveness and necessity of these enhancements in refining the model's segmentation accuracy.

In such a way, Table~\ref{tab:tribert_ablation}, which presents the TriBERT dataset, follows the same trend as in M4GT dataset, where the TNC-DBGC alone has given a overall result of 0.737, however upon the incremental addition of these optimization techniques have given us the value of 0.806 in the fully optimized TNC-DBGC\textsuperscript{*} configuration.

\section{Conclusion and Future Work}
The widespread availability of LLMs has introduced significant challenges across multiple domains, where controlling the inappropriate or unintended use of these models has become increasingly difficult. To overcome these problems, in this paper, we propose an approach to segment the human and AI spans in a collaboratively written text. Our approach uses a hybrid component architecture that combines 1) Transformers, 2) Neural Networks, and 3) CRF. For the justification to prove our model performs better than existing ones, we evaluated our method on two Human-AI collaborative text datasets: 1) TriBERT and 2) M4GT, and achieved state-of-the-art results on all of them, and compared the results with zero-shot approaches that include FastDetectGPT, Glimpse, and Binoculars, and also performed the experimentation with HMMs and MEMMs. 

As a future work, we wanted to enhance our model by incorporating style features, such as lexical diversity or syntactic complexity, to improve boundary detection even under adversarial conditions, syntactical or semantical ones. Exploring multi-task learning to jointly predict authorship and boundary points may further improve the performance. Additionally, evaluating the model on diverse datasets with various AI-generated texts and testing its generalizability across languages could broaden its applicability.

\section{Limitations}
Although our models and methodologies achieved promising results, a key limitation is the lack of robustness against both syntactic and semantic adversarial attacks. The model has not been explicitly trained or tested on adversarial samples, which could potentially manipulate linguistic structures or semantics to mislead sequence labeling. Addressing these challenges in future work could significantly enhance model reliability and generalization in real-world applications.

\bibliography{custom}   

\appendix

\section{Details on Datasets}
\label{appendix_datasets}
Table~\ref{tab:dataset1} tells about the TriBERT \cite{zeng2024towards} Hybrid Mixed Text Dataset statistics, which details the distribution of academic essays with 1, 2, or 3 human-AI authorship boundaries. It includes metrics such as essay counts (\textit{17,136} total), average words (\textit{287.6}) and sentences (\textit{13.7}) per essay, average lengths of AI-generated (\textit{22.2} words) and human-written (\textit{22.4} words) sentences, and at last the proportion of AI-generated sentences (65.3\%).

\noindent Table~\ref{tab:dataset2} gives an overview about the M4GT Boundary Identification Dataset \cite{wang2024m4gt}, which includes the PeerRead and OUTFOX subsets. PeerRead includes \textit{5,676} samples each for ChatGPT and LLaMA-2 models (\textit{7B, 13B, 70B}), split into train (\textit{3,649}), dev (\textit{505}), and test (\textit{1,522}) sets, with an additional \textit{5,189} samples for LLaMA-2-7B*. OUTFOX provides \textit{1,000} test samples each for GPT-4 and LLaMA-2 variants.

\begin{table}[htbp]
\scriptsize
\centering
\resizebox{1.0\linewidth}{!}{%
\begin{tabular}{lcccc}
\toprule
 & \multicolumn{3}{c}{\textbf{Boundaries}} & \textbf{All} \\
\cmidrule(lr){2-4}
 & \textbf{1} & \textbf{2} & \textbf{3} & \\
\midrule
Hybrid essay & 7488 & 6429 & 3219 & 17136 \\
Words per essay & 275.3 & 279.5 & 332.6 & 287.6 \\
Sentences per essay & 12.9 & 13.4 & 16.1 & 13.7 \\
Avg len of AI-gen sent & 22.7 & 21.8 & 21.7 & 22.2 \\
Avg len of human-written sent & 22.7 & 22.6 & 21.2 & 22.4 \\
Ratio of AI-gen sent per essay & 67.4\% & 58.8\% & 73.2\% & 65.3\% \\
\bottomrule
\end{tabular}
}
\caption{TriBERT Hybrid Mixed text Dataset statistics by \cite{zeng2024towards}}
\label{tab:dataset1}
\end{table}

\begin{table}[ht]
\scriptsize
\centering
\resizebox{1.0\linewidth}{!}{%
\begin{tabular}{ll|cccc}
\hline
\textbf{Domain} & \textbf{Generator} & \textbf{Train} & \textbf{Dev} & \textbf{Test} & \textbf{Total} \\ \hline
\multirow{5}{*}{PeerRead} 
& ChatGPT          & 3,649 & 505 & 1,522 & 5,676\\ 
& LLaMA-2-7B*      & 3,649 & 505 & 1,035 & 5,189\\ 
& LLaMA-2-7B       & 3,649 & 505 & 1,522 & 5,676\\ 
& LLaMA-2-13B      & 3,649 & 505 & 1,522 & 5,676\\ 
& LLaMA-2-70B      & 3,649 & 505 & 1,522 & 5,676\\ \hline
\multirow{4}{*}{OUTFOX} 
& GPT-4            & –           & –         & 1,000 & 1,000\\ 
& LLaMA-2-7B       & –           & –         & 1,000 & 1,000\\ 
& LLaMA-2-13B      & –           & –         & 1,000 & 1,000\\ 
& LLaMA-2-70B      & –           & –         & 1,000 & 1,000\\ \hline
\end{tabular}
}
\caption{M4GT Boundary identification data based on GPT and LLaMA-2 series \cite{wang2024m4gt}.}
\label{tab:dataset2}
\end{table}

\section{Other Experimental Results}
As mentioned above, other than the main evaluation metrics, we have also computed the Accuracy, Precision, Recall, F1-Score, MCC, and Cohen's Kappa score for both datasets. All best model in each setting are highlighted in bold. According to the metrics, in \textit{NN\_CRF} setting, the BiGRU\_CRF model has the highest scores among the other models. In \textit{Transformer\_CRF} setting, DeBERTa\_CRF model got the highest accuracy with 97.82\%. Finally, among the \textit{Transformer\_NN\_CRF} setting, the model DeBERTa\_BiGRU\_CRF has the highest scores in all the metrics and outperforms all the models, including the zero-shot and the models that were previously tested on. The results of these metrics can be seen in the Tables \ref{tab:master_results_AAAI} and \ref{tab:master_results_M4GT}.

\begin{figure}[htbp]
  \centering
  \begin{subfigure}[b]{0.40\textwidth}
    \centering
    \includegraphics[width=\linewidth]{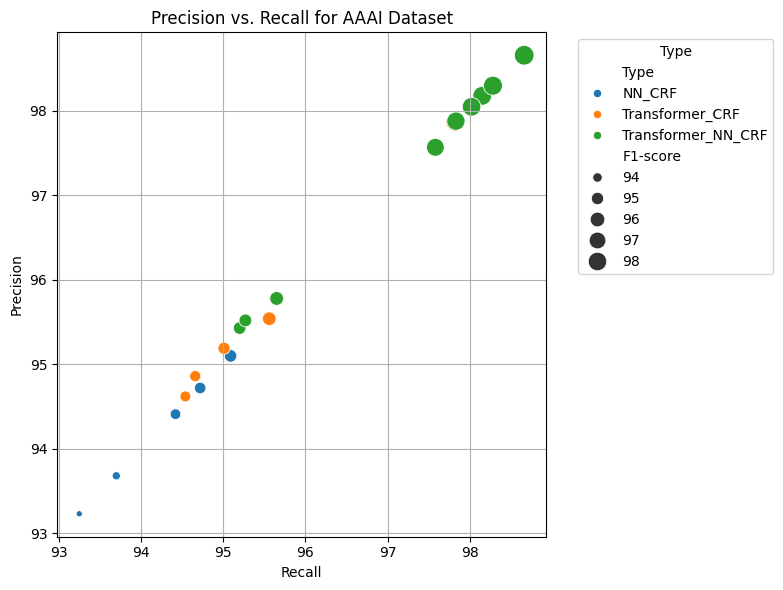}
    \caption{TriBERT Precision vs Recall}
    \label{fig:tribert_precision_recall}
  \end{subfigure}
  \hfill
  \begin{subfigure}[b]{0.40\textwidth}
    \centering
    \includegraphics[width=\linewidth]{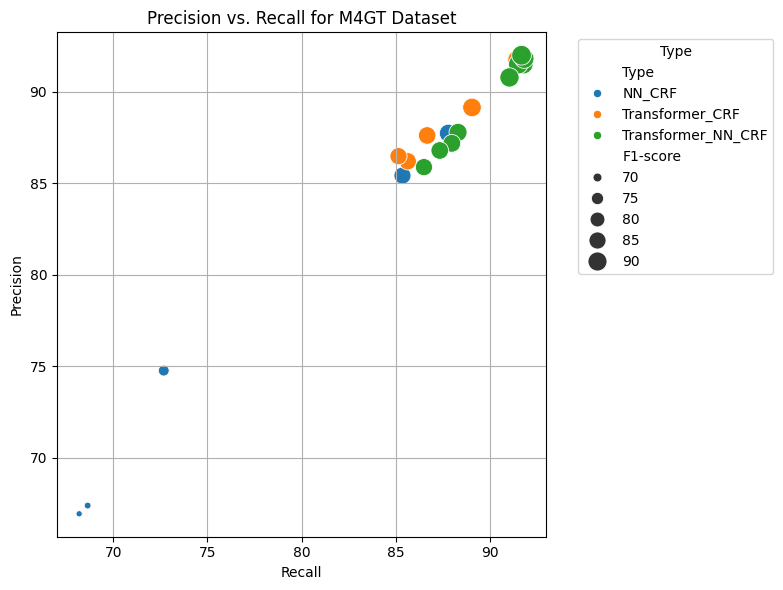}
    \caption{M4GT Precision vs Recall}
    \label{fig:m4gt_precision_recall}
  \end{subfigure}
  \caption{Precision vs Recall plots for TriBERT and M4GT datasets.}
  \label{fig:precision_recall}
\end{figure}

\begin{figure}[htbp]
  \centering
    \includegraphics[width=0.65\linewidth]{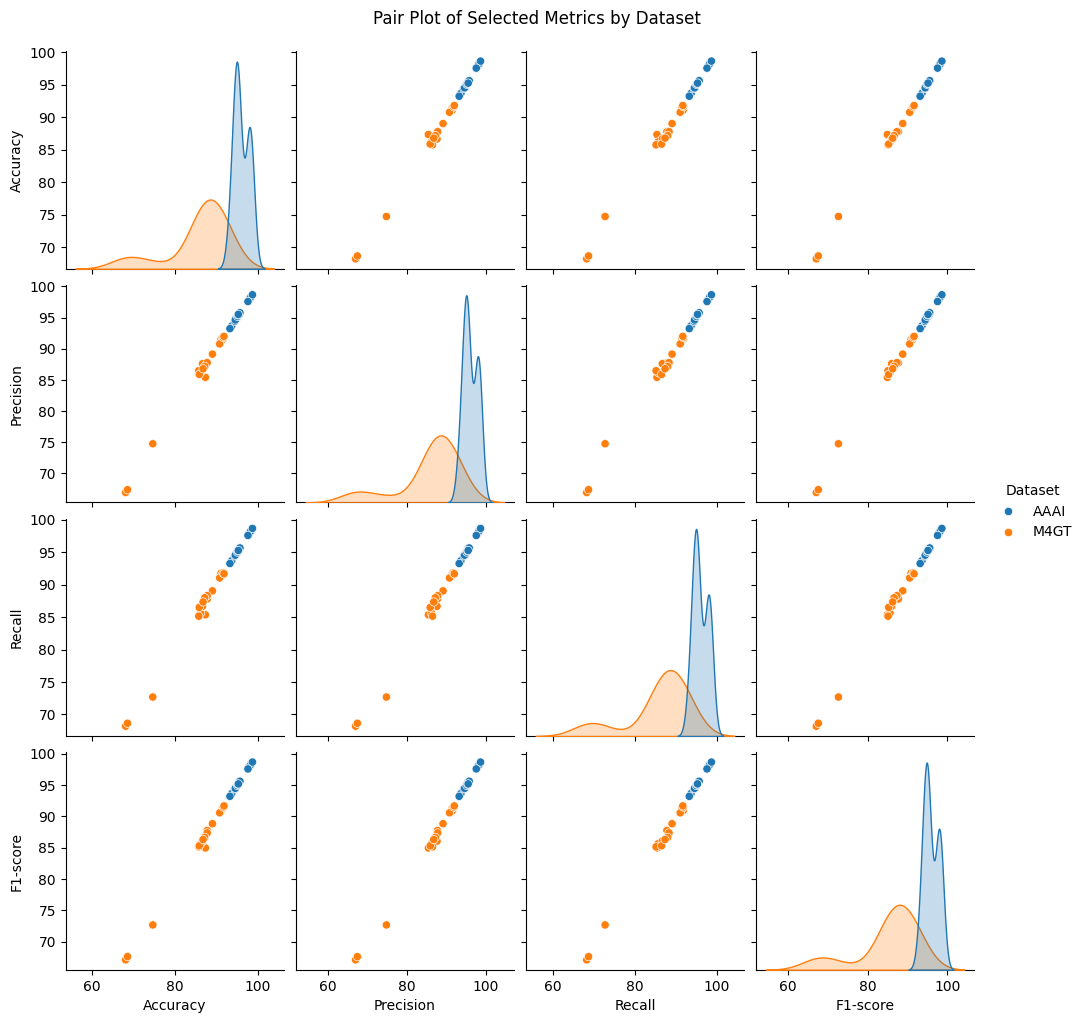}
    \caption{Combined Result values over the two Datasets.}
    \label{comnied_results}
\end{figure}

\begin{figure}[htbp]
  \centering
  \begin{subfigure}[b]{0.40\textwidth}
    \centering
    \includegraphics[width=\linewidth]{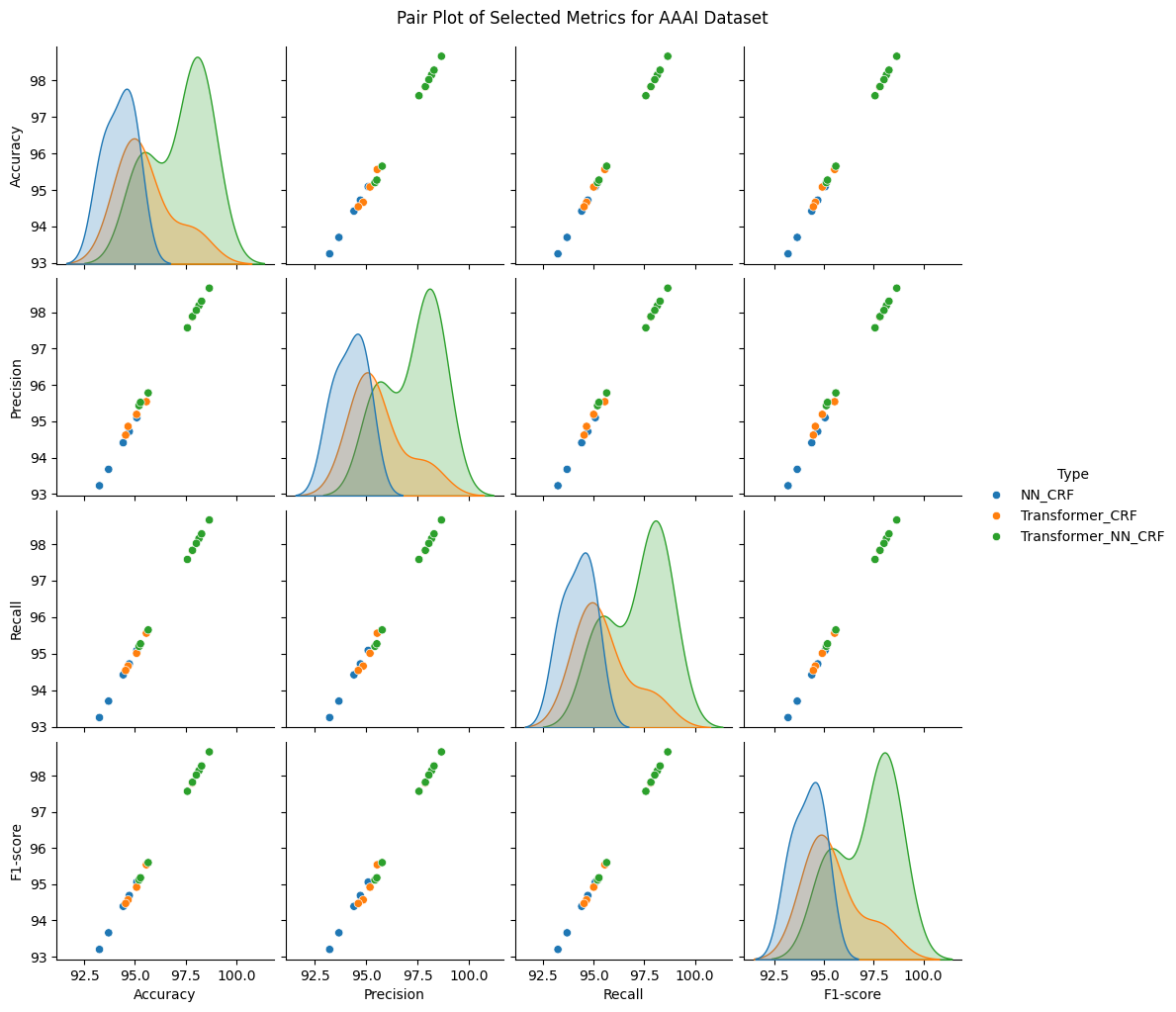}
    \caption{TriBERT Pair Plot}
    \label{fig:tribert_pair_plot}
  \end{subfigure}
  \hfill
  \begin{subfigure}[b]{0.40\textwidth}
    \centering
    \includegraphics[width=\linewidth]{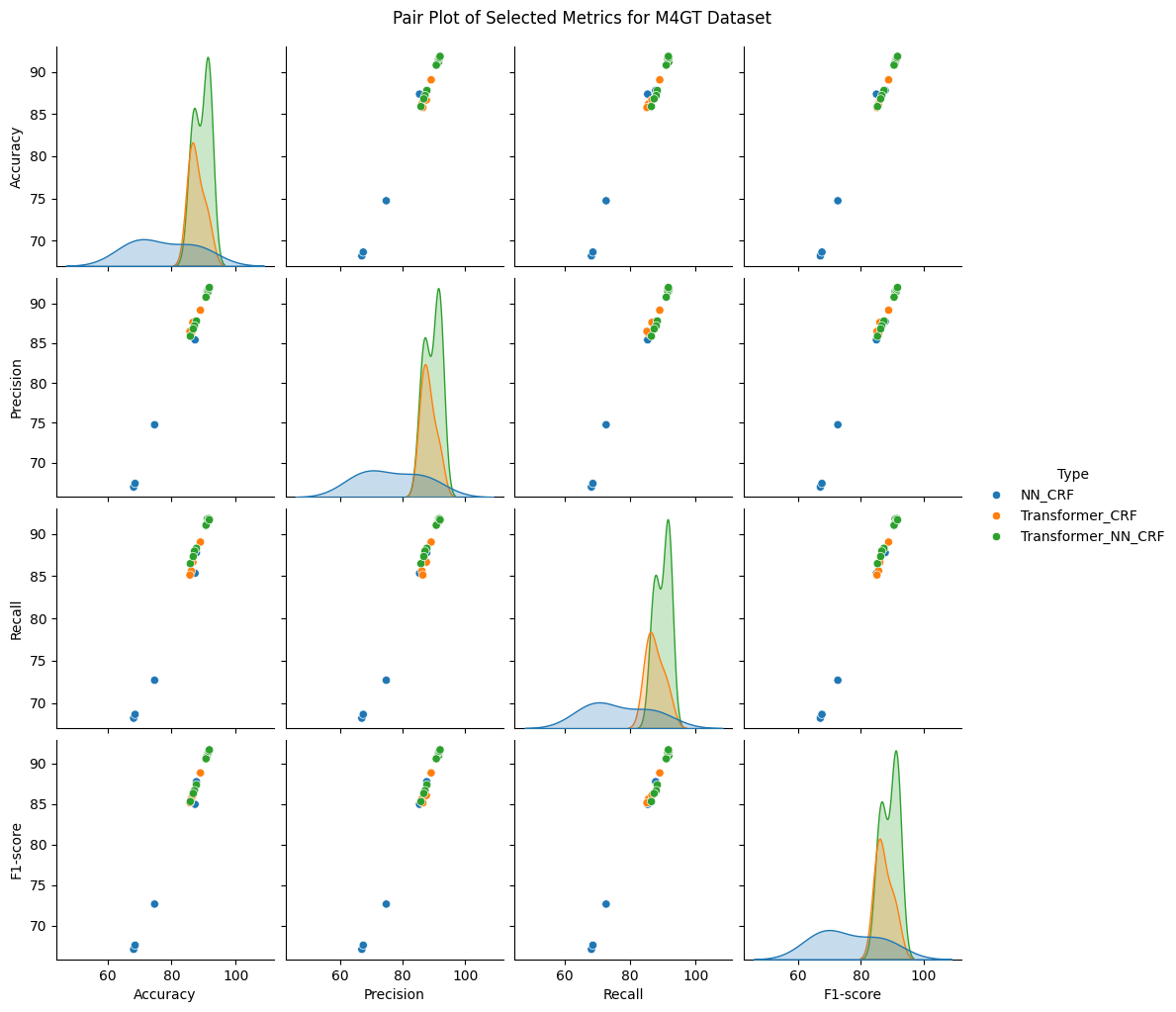}
    \caption{M4GT Pair Plot}
    \label{fig:m4gt_pair_plot}
  \end{subfigure}
  \caption{Pair plots for TriBERT and M4GT datasets.}
  \label{fig:pair_plots}
\end{figure}

\begin{figure}[htbp]
  \centering
  \begin{subfigure}[b]{0.40\textwidth}
    \centering
    \includegraphics[width=\linewidth]{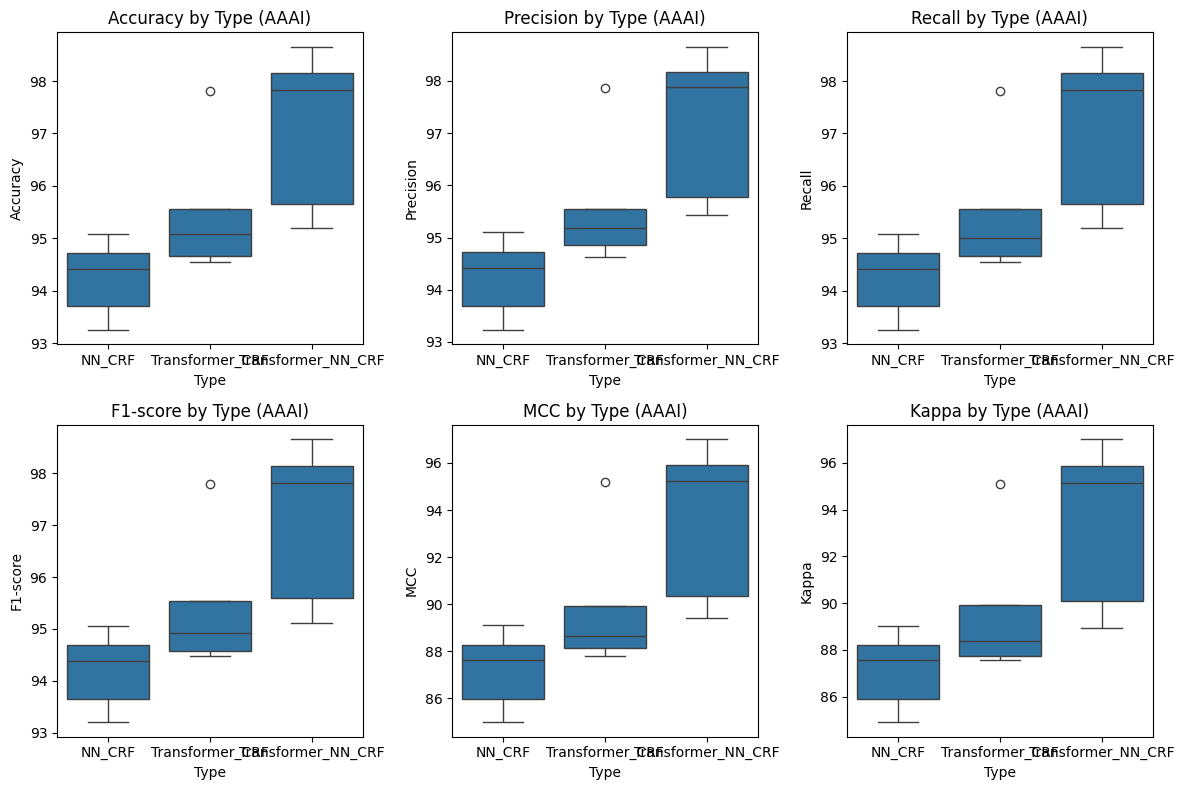}
    \caption{TriBERT Box Plot}
    \label{fig:tribert_box_plot}
  \end{subfigure}
  \hfill
  \begin{subfigure}[b]{0.40\textwidth}
    \centering
    \includegraphics[width=\linewidth]{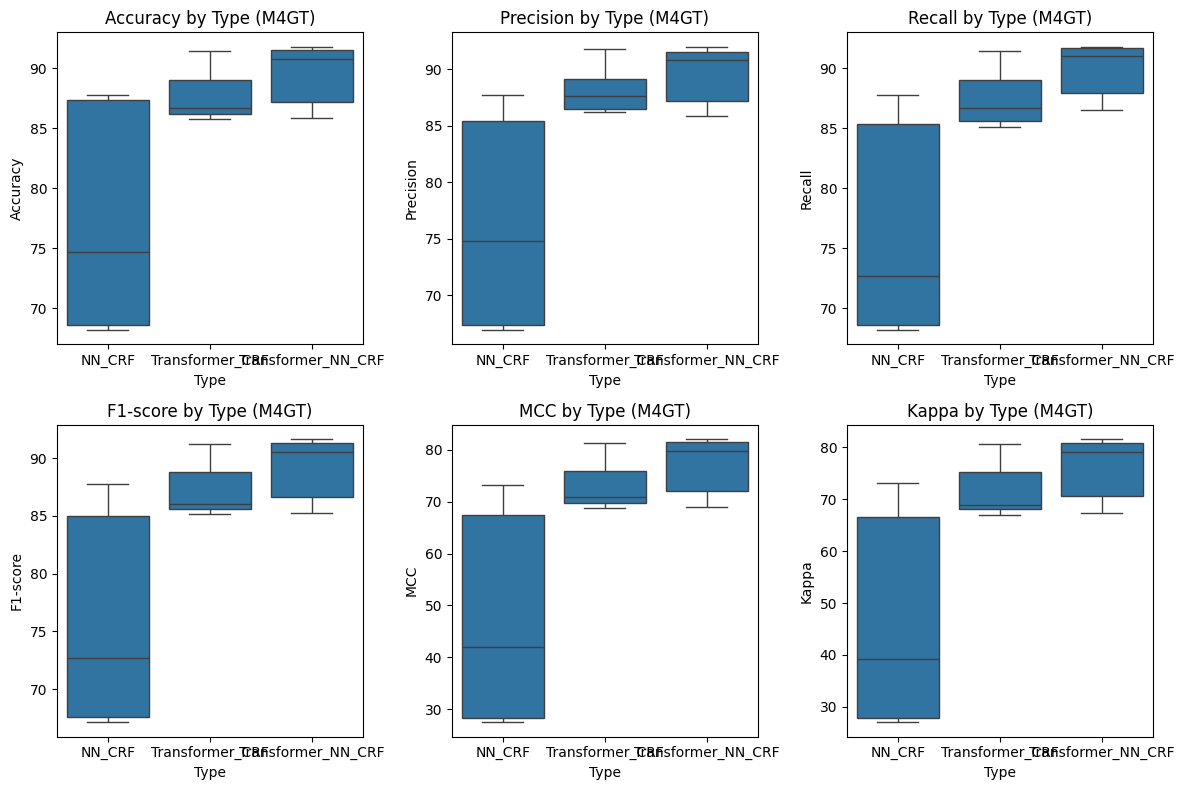}
    \caption{M4GT Box Plot}
    \label{fig:m4gt_box_plot}
  \end{subfigure}
  \caption{Box plots for TriBERT and M4GT datasets.}
  \label{fig:box_plots}
\end{figure}

According to figures \ref{fig:precision_recall}, \ref{fig:box_plots}, \ref{fig:pair_plots}, it is certain that the increase in the combination of number of blocks in a model leads to have high values in the evaluation metrics and were able to outperform the other models with their previous setting. In conclusion, the results from the TriBERT Dataset are quite high, and the results on the M4GT dataset are comparatively low. This shows that the robustness of the data is higher in the M4GT dataset, where a model cannot easily predict the boundary in the M4GT dataset and can predict easily in the TriBERT dataset. The results comparison plot for both datasets can be seen in Figure~\ref{comnied_results}.

\begin{table*}[h!]
\centering
\scriptsize
\resizebox{1.0\linewidth}{!}{%
\begin{tabular}{@{}l|l|c|l|l|l|l|l|l|l@{}}
\toprule
\textbf{Dataset} & \textbf{Type} & \textbf{Model} & \textbf{Accuracy} & \textbf{Precision} & \textbf{Recall} & \textbf{F1-score} & \textbf{MCC} & \textbf{Kappa} \\
\midrule
\multirow{19}{*}{AAAI} & \multirow{5}{*}{NN\_CRF} & CNN\_CRF & 94.72 & 94.72 & 94.72 & 94.69 & 88.28 & 88.22 \\
& & RNN\_CRF & 93.7 & 93.68 & 93.7 & 93.66 & 85.98 & 85.92 \\
& & LSTM\_CRF & 93.25 & 93.23 & 93.25 & 93.2 & 84.97 & 84.9 \\
& & BiLSTM\_CRF & 94.42 & 94.41 & 94.42 & 94.39 & 87.61 & 87.55 \\
& & \textbf{BiGRU\_CRF} & \textbf{95.09} & \textbf{95.1} & \textbf{95.09} & \textbf{95.06} & \textbf{89.1} & \textbf{89.04} \\
\cmidrule(l){2-9}
& \multirow{5}{*}{Transformer\_CRF} & BERT\_CRF & 94.66 & 94.86 & 94.66 & 94.57 & 88.15 & 87.76 \\
& & DistilBERT\_CRF & 94.54 & 94.62 & 94.54 & 94.47 & 87.8 & 87.57 \\
& & RoBERTa\_CRF & 95.08 & 95.19 & 95.01 & 94.92 & 88.65 & 88.37 \\
& & ModernBERT\_CRF & 95.56 & 95.54 & 95.56 & 95.54 & 89.92 & 89.9 \\
& & \textbf{DeBERTa\_CRF} & \textbf{97.82} & \textbf{97.87} & \textbf{97.82} & \textbf{97.8} & \textbf{95.18} & \textbf{95.09} \\
\cmidrule(l){2-9}
& \multirow{9}{*}{Transformer\_NN\_CRF} & DeBERTa\_CNN\_CRF & 98.15 & 98.18 & 98.15 & 98.14 & 95.91 & 95.86 \\
& & DeBERTa\_RNN\_CRF & 97.83 & 97.88 & 97.83 & 97.82 & 95.22 & 95.13 \\
& & DeBERTa\_LSTM\_CRF & 98.02 & 98.05 & 98.02 & 98 & 95.62 & 95.55 \\
& & DeBERTa\_BiLSTM\_CRF & 98.28 & 98.3 & 98.28 & 98.27 & 96.2 & 96.16 \\
& & \textbf{DeBERTa\_BiGRU\_CRF} & \textbf{98.66} & \textbf{98.66} & \textbf{98.66} & \textbf{98.66} & \textbf{97.02} & \textbf{97.02} \\
& & BERT\_BiGRU\_CRF & 95.65 & 95.78 & 95.65 & 95.6 & 90.35 & 90.09 \\
& & DistilBERT\_BiGRU\_CRF & 95.2 & 95.43 & 95.2 & 95.12 & 89.49 & 88.99 \\
& & RoBERTa\_BiGRU\_CRF & 95.27 & 95.52 & 95.27 & 95.18 & 89.41 & 88.95 \\
& & ModernBERT\_BiGRU\_CRF & 97.58 & 97.57 & 97.58 & 97.57 & 94.51 & 94.49 \\
\bottomrule
\end{tabular}
}
\caption{Conventional metrics results table that included the metrics Accuracy, Precision, Recall, F1-Score, MCC and Kappa score on the AAAI-TriBERT dataset by \citet{zeng2024towards}.}
\label{tab:master_results_AAAI}
\end{table*}

\begin{table*}[h!]
\centering
\scriptsize
\resizebox{1.0\linewidth}{!}{%
\begin{tabular}{@{}l|l|c|l|l|l|l|l|l|l@{}}
\toprule
\textbf{Dataset} & \textbf{Type} & \textbf{Model} & \textbf{Accuracy} & \textbf{Precision} & \textbf{Recall} & \textbf{F1-score} & \textbf{MCC} & \textbf{Kappa} \\
\midrule
\multirow{19}{*}{M4GT} & \multirow{5}{*}{NN\_CRF} & CNN\_CRF & 74.72 & 74.77 & 72.68 & 72.68 & 41.98 & 39.18 \\
& & RNN\_CRF & 68.19 & 66.95 & 68.19 & 67.11 & 27.45 & 26.98 \\
& & LSTM\_CRF & 68.64 & 67.4 & 68.64 & 67.62 & 28.39 & 27.87 \\
& & BiLSTM\_CRF & 87.35 & 85.42 & 85.35 & 84.96 & 67.43 & 66.67 \\
& & \textbf{BiGRU\_CRF} & \textbf{87.79} & \textbf{87.73} & \textbf{87.79} & \textbf{87.75} & \textbf{73.24} & \textbf{73.22} \\
\cmidrule(l){2-9}
& \multirow{5}{*}{Transformer\_CRF} & BERT\_CRF & 86.2 & 86.2 & 85.63 & 85.63 & 69.67 & 68.04 \\
& & DistilBERT\_CRF & 89.04 & 89.14 & 89.04 & 88.83 & 75.82 & 75.32 \\
& & RoBERTa\_CRF & 86.66 & 87.61 & 86.66 & 86.05 & 70.94 & 68.96 \\
& & ModernBERT\_CRF & 85.76 & 86.48 & 85.14 & 85.14 & 68.71 & 66.94 \\
& & \textbf{DeBERTa\_CRF} & \textbf{91.41} & \textbf{91.75} & \textbf{91.46} & \textbf{91.22} & \textbf{81.28} & \textbf{80.57} \\
\cmidrule(l){2-9}
& \multirow{9}{*}{Transformer\_NN\_CRF} & DeBERTa\_CNN\_CRF & 91.13 & 91.5 & 91.77 & 90.92 & 80.65 & 79.92 \\
& & DeBERTa\_RNN\_CRF & 91.5 & 91.5 & 91.5 & 91.32 & 81.43 & 80.82 \\
& & DeBERTa\_LSTM\_CRF & 90.78 & 90.78 & 91.03 & 90.57 & 79.8 & 79.15 \\
& & DeBERTa\_BiLSTM\_CRF & 91.64 & 91.81 & 91.81 & 91.49 & 81.68 & 81.22 \\
& & \textbf{DeBERTa\_BiGRU\_CRF} & \textbf{91.82} & \textbf{91.99} & \textbf{91.67} & \textbf{91.67} & \textbf{82.08} & \textbf{81.62} \\
& & BERT\_BiGRU\_CRF & 85.88 & 85.88 & 86.49 & 85.3 & 68.9 & 67.31 \\
& & DistilBERT\_BiGRU\_CRF & 87.78 & 87.78 & 88.3 & 87.36 & 73.21 & 71.94 \\
& & RoBERTa\_BiGRU\_CRF & 87.18 & 87.18 & 87.97 & 86.65 & 72.02 & 70.31 \\
& & ModernBERT\_BiGRU\_CRF & 86.79 & 86.79 & 87.33 & 86.3 & 70.96 & 70.57 \\
\bottomrule
\end{tabular}
}
\caption{Conventional metrics results table that included the metrics Accuracy, Precision, Recall, F1-Score, MCC, and Kappa score on the M4GT dataset by \citet{wang2024m4gt}.}
\label{tab:master_results_M4GT}
\end{table*}

\clearpage
\section{Loss Plots}
\begin{figure}[htbp]
  \centering
  \begin{subfigure}[b]{0.45\textwidth}
    \centering
    \includegraphics[width=\linewidth]{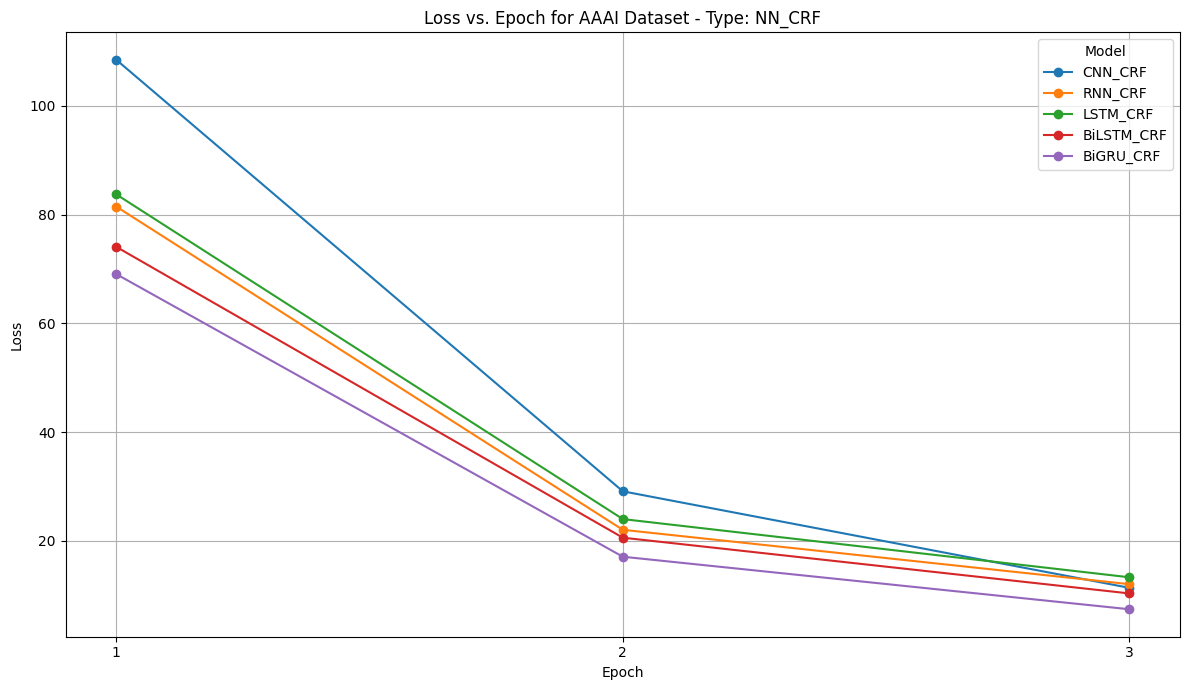}
    \label{fig:tribert_nn_crf_loss}
  \end{subfigure}
  \hfill
  \begin{subfigure}[b]{0.45\textwidth}
    \centering
    \includegraphics[width=\linewidth]{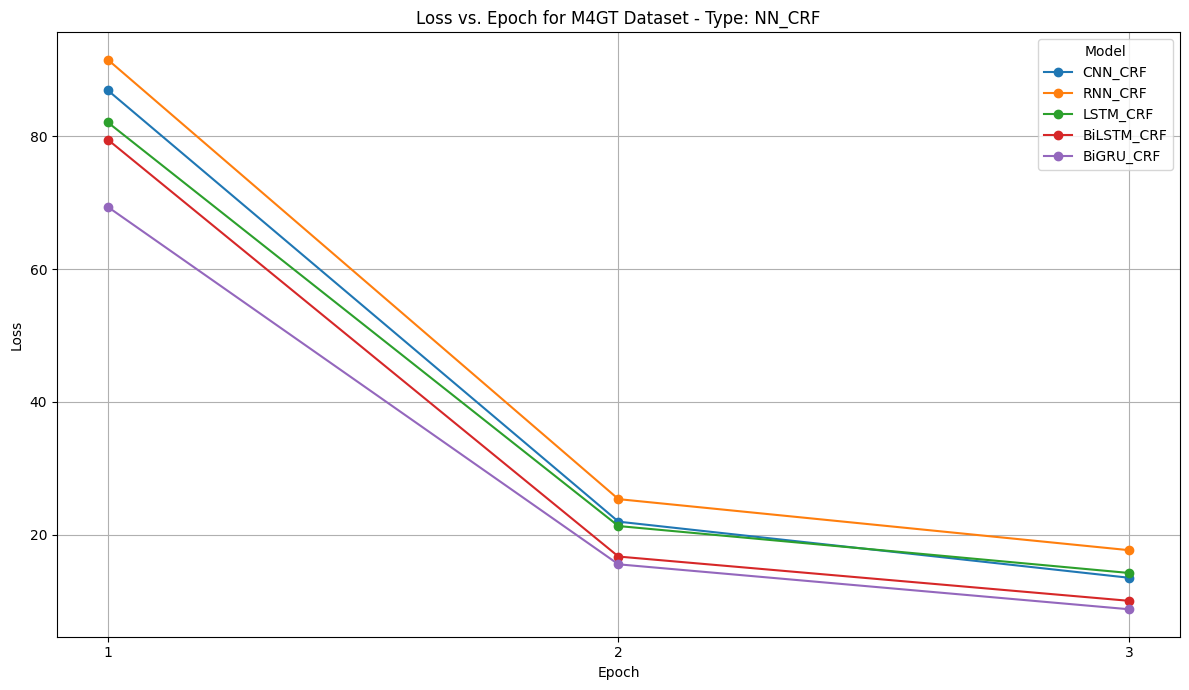}
    \label{fig:m4gt_nn_crf_loss}
  \end{subfigure}
  \caption{Loss curves for neural network with CRF models on TriBERT and M4GT datasets.}
  \label{fig:nn_crf_loss_plots}
\end{figure}

\begin{figure}[htbp]
  \centering
  \begin{subfigure}[b]{0.45\textwidth}
    \centering
    \includegraphics[width=\linewidth]{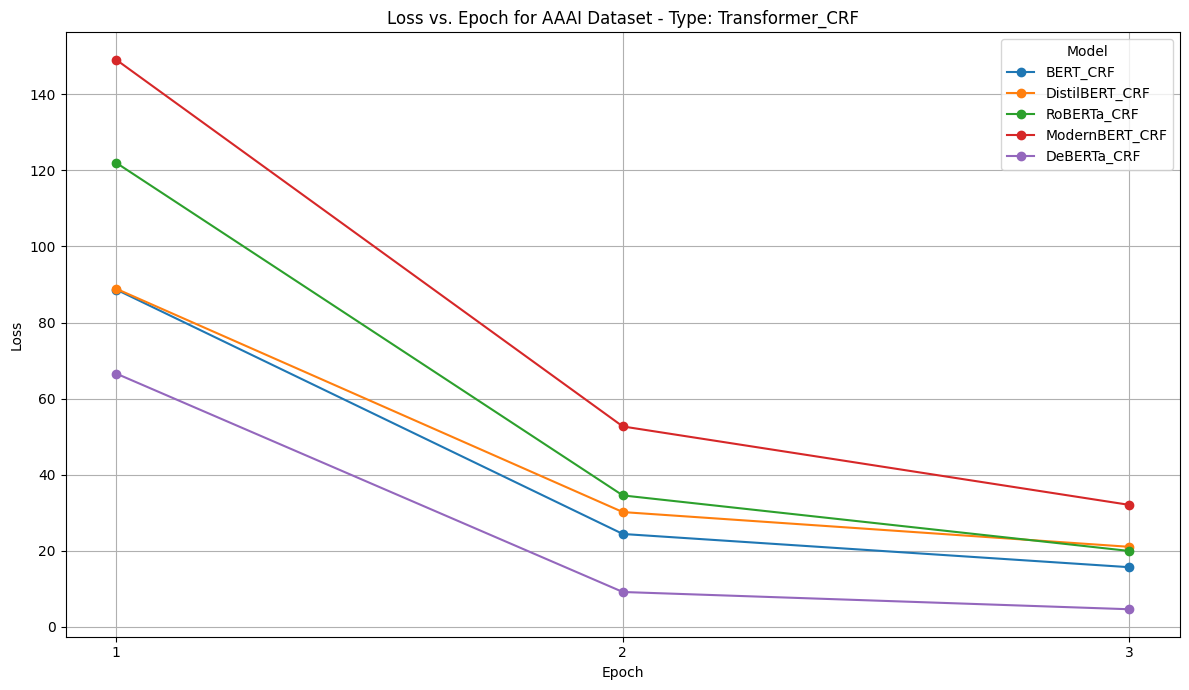}
    \label{fig:tribert_t_crf_loss}
  \end{subfigure}
  \hfill
  \begin{subfigure}[b]{0.45\textwidth}
    \centering
    \includegraphics[width=\linewidth]{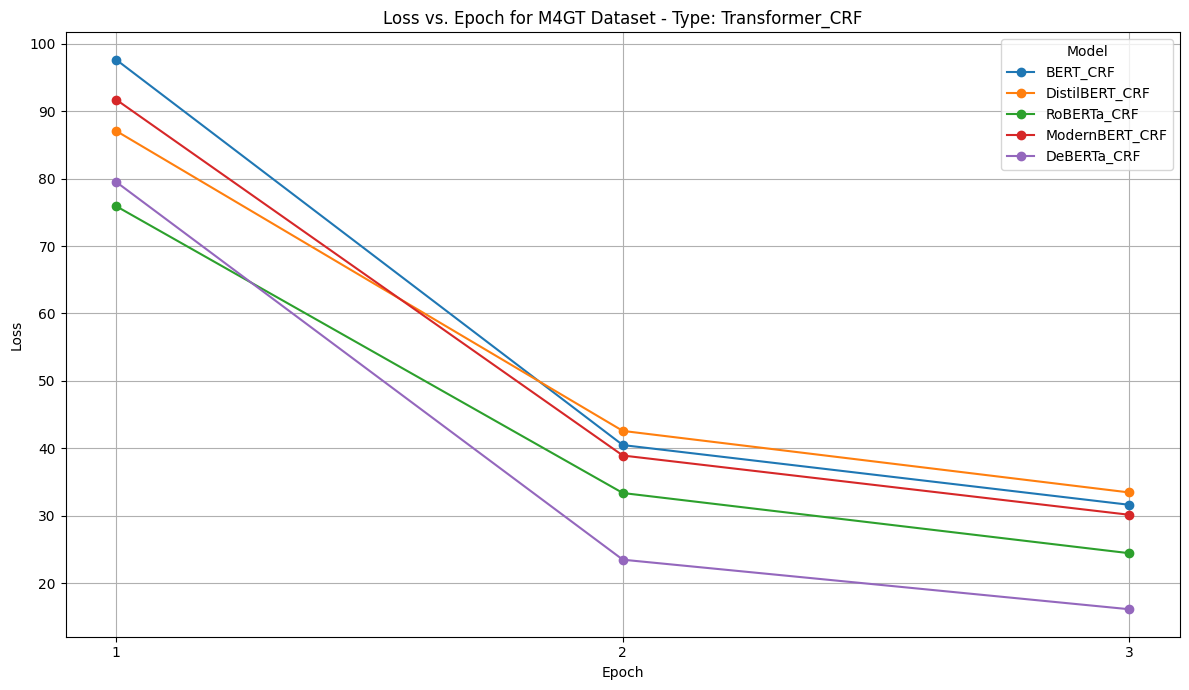}
    \label{fig:m4gt_t_crf_loss}
  \end{subfigure}
  \caption{Loss curves for transformer with CRF models on TriBERT and M4GT datasets.}
  \label{fig:t_crf_loss_plots}
\end{figure}

\begin{figure}[htbp]
  \centering
  \begin{subfigure}[b]{0.45\textwidth}
    \centering
    \includegraphics[width=\linewidth]{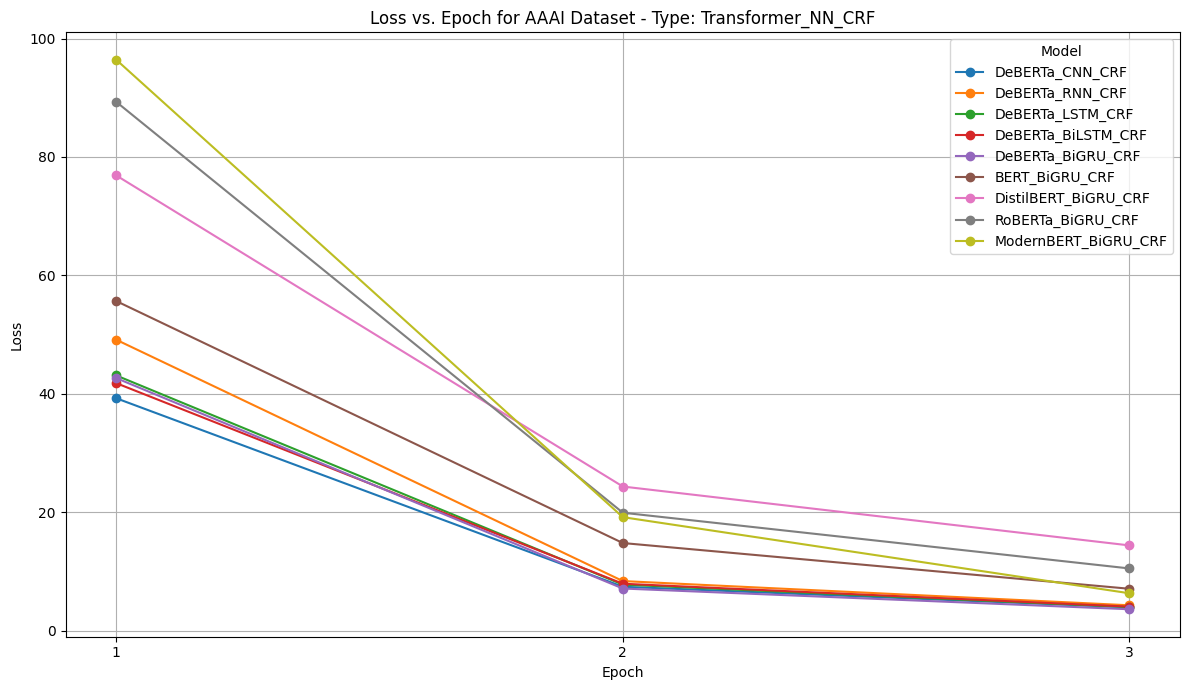}
    \label{fig:tribert_t_nn_crf_loss}
  \end{subfigure}
  \hfill
  \begin{subfigure}[b]{0.45\textwidth}
    \centering
    \includegraphics[width=\linewidth]{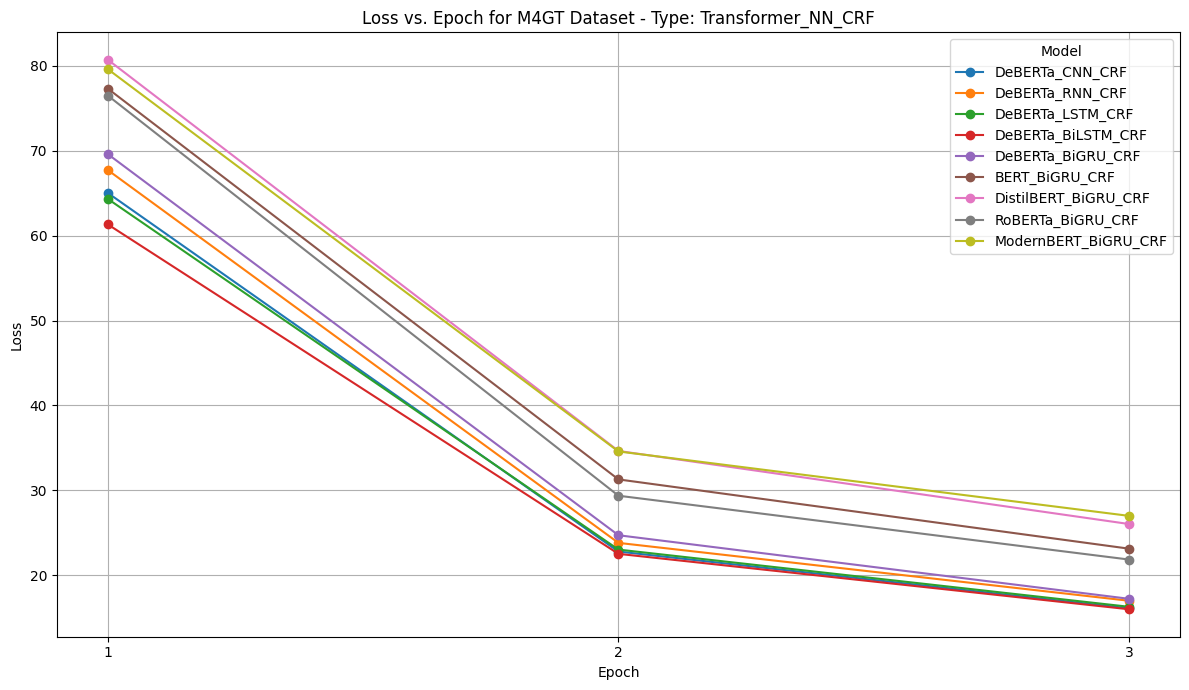}
    \label{fig:m4gt_t_nn_crf_loss}
  \end{subfigure}
  \caption{Loss curves for transformer and neural network with CRF models on TriBERT and M4GT datasets.}
  \label{fig:t_nn_crf_loss_plots}
\end{figure}

Our model includes the CRF layers such that the loss during the training of the model is the CRF loss, where CRF tries to maximize the log-likelihood of the correct label sequence. Such that the loss at the first epochs would be very high, and upon predicting the token-label correctly, the loss would eventually decrease. The numerical values during the training of the models on two datasets can be seen in the Tables \ref{tab:loss_epochs_AAAI}, \ref{tab:loss_epochs_M4GT}. A better visualization of the decrease over the training epochs can be seen in the Figures \ref{fig:nn_crf_loss_plots}, \ref{fig:t_crf_loss_plots}, and \ref{fig:t_nn_crf_loss_plots}.

The CRF loss is mathematically expressed as:
\begin{equation}
\mathcal{L}_{CRF} = -\log P(y \mid x)
\end{equation}
where: \(P(y \mid x)\) is the conditional probability and is given in the equation \ref{eq:pyx_prob}.
This is further expanded as follows when Equation \ref{eq:pyx_prob} is substituted above:
\begin{equation}
\mathcal{L}_{CRF} = -\left(S(x, y) - \log Z(x)\right)
\end{equation}
Where: \(S(x, y)\) is the sum of the transition scores and emission scores of the CRF model, and the equation is \ref{crf_scorer}. \(\log Z(x)\) is the log of the partition function \(Z(x)\) and the Equation is \ref{eq:zofx}. The loss minimization reflects the CRF's ability to capture dependencies between labels, improving boundary detection accuracy over epochs.

\begin{table*}[h!]
\centering
\scriptsize
\begin{tabular}{@{}l|l|c|l|l|ll@{}}
\toprule
\textbf{Dataset} & \textbf{Type} & \textbf{Model} & \textbf{Epoch 1} & \textbf{Epoch 2} & \textbf{Epoch 3} \\
\midrule
\multirow{19}{*}{AAAI} & \multirow{5}{*}{NN\_CRF} & CNN\_CRF & 108.47 & 29.12 & 11.37 \\
& & RNN\_CRF & 81.48 & 22.04 & 12.04 \\
& & LSTM\_CRF & 83.72 & 24.01 & 13.31 \\
& & BiLSTM\_CRF & 74.07 & 20.59 & 10.34 \\
& & \textbf{BiGRU\_CRF} & 69.06 & 17.09 & 7.43 \\
\cmidrule(l){2-6}
& \multirow{5}{*}{Transformer\_CRF} & BERT\_CRF & 88.66 & 24.38 & 15.64 \\
& & DistilBERT\_CRF & 88.87 & 30.14 & 21 \\
& & RoBERTa\_CRF & 121.99 & 34.52 & 19.91 \\
& & ModernBERT\_CRF & 149.1 & 52.66 & 32.03 \\
& & \textbf{DeBERTa\_CRF} & 66.56 & 9.12 & 4.58 \\
\cmidrule(l){2-6}
& \multirow{9}{*}{Transformer\_NN\_CRF} & DeBERTa\_CNN\_CRF & 39.26 & 7.44 & 3.96 \\
& & DeBERTa\_RNN\_CRF & 49.14 & 8.4 & 4.31 \\
& & DeBERTa\_LSTM\_CRF & 43.11 & 7.8 & 4 \\
& & \textbf{DeBERTa\_BiLSTM\_CRF} & 41.81 & 7.95 & 4.08 \\
& & DeBERTa\_BiGRU\_CRF & 42.66 & 7.14 & 3.64 \\
& & BERT\_BiGRU\_CRF & 55.67 & 14.82 & 7.1 \\
& & DistilBERT\_BiGRU\_CRF & 76.88 & 24.36 & 14.42 \\
& & RoBERTa\_BiGRU\_CRF & 89.3 & 19.94 & 10.53 \\
& & ModernBERT\_BiGRU\_CRF & 96.41 & 19.19 & 6.34 \\
\bottomrule
\end{tabular}
\caption{Model Loss Across Epochs during training of the model on the TriBERT dataset.}
\label{tab:loss_epochs_AAAI}
\end{table*}

\begin{table*}[h!]
\centering
\scriptsize
\begin{tabular}{@{}l|l|c|l|l|ll@{}}
\toprule
\textbf{Dataset} & \textbf{Type} & \textbf{Model} & \textbf{Epoch 1} & \textbf{Epoch 2} & \textbf{Epoch 3} \\
\midrule
\multirow{19}{*}{M4GT} & \multirow{5}{*}{NN\_CRF} & CNN\_CRF & 86.87 & 21.98 & 13.54 \\
& & RNN\_CRF & 91.48 & 25.37 & 17.69 \\
& & LSTM\_CRF & 82.05 & 21.32 & 14.26 \\
& & BiLSTM\_CRF & 79.45 & 16.73 & 10.07 \\
& & \textbf{BiGRU\_CRF} & 69.33 & 15.56 & 8.8 \\
\cmidrule(l){2-6}
& \multirow{5}{*}{Transformer\_CRF} & BERT\_CRF & 97.62 & 40.48 & 31.62 \\
& & DistilBERT\_CRF & 87.06 & 42.58 & 33.45 \\
& & \textbf{RoBERTa\_CRF} & 75.9 & 33.37 & 24.44 \\
& & ModernBERT\_CRF & 91.7 & 38.93 & 30.14 \\
& & DeBERTa\_CRF & 79.49 & 23.48 & 16.13 \\
\cmidrule(l){2-6}
& \multirow{9}{*}{Transformer\_NN\_CRF} & DeBERTa\_CNN\_CRF & 65 & 22.81 & 16.13 \\
& & DeBERTa\_RNN\_CRF & 67.69 & 23.83 & 17.02 \\
& & DeBERTa\_LSTM\_CRF & 64.3 & 23.04 & 16.31 \\
& & \textbf{DeBERTa\_BiLSTM\_CRF} & 61.3 & 22.54 & 16.02 \\
& & DeBERTa\_BiGRU\_CRF & 69.57 & 24.73 & 17.25 \\
& & BERT\_BiGRU\_CRF & 77.28 & 31.3 & 23.15 \\
& & DistilBERT\_BiGRU\_CRF & 80.68 & 34.65 & 26.05 \\
& & RoBERTa\_BiGRU\_CRF & 76.46 & 29.38 & 21.87 \\
& & ModernBERT\_BiGRU\_CRF & 79.57 & 34.59 & 27.01 \\
\bottomrule
\end{tabular}
\caption{Model Loss Across Epochs during training of the model on the M4GT dataset}
\label{tab:loss_epochs_M4GT}
\end{table*}

\end{document}